\documentclass[preprint,12pt,authoryear]{elsarticle}

\usepackage{amssymb}
\usepackage{amsmath}
\usepackage{color}
\usepackage{algorithmic}
\usepackage[linesnumbered,ruled]{algorithm2e}
\usepackage{threeparttable}
\usepackage{multirow}
\usepackage{graphicx}
\usepackage {verbatim}
\makeatletter \renewcommand{\@biblabel}[1]{} \makeatother
\usepackage[hidelinks]{hyperref}
\hypersetup{
    colorlinks=True
}

\journal{Medical Image Analysis}

\begin{document}

\begin{frontmatter}



\title{Diff-CL: A Novel Cross Pseudo-Supervision Method for Semi-supervised Medical Image Segmentation}


\author[1]{Xiuzhen Guo} 
\ead{xiuzhenguo589@gmail.com}
\author[1]{Lianyuan Yu}
\ead{lianyuanyu8@gmail.com}
\author[2]{Ji Shi \corref{cor1}}
\ead{shiji@cnu.edu.cn}
\cortext[cor1]{Corresponding authors.}
\author[3]{Na Lei \corref{cor1}}
\ead{nalei@dlut.edu.cn}
\author[2]{Hongxiao Wang}
\ead{hwang21@cnu.edu.cn}
\affiliation[1]{organization={School of Mathematical Science},
            addressline={Capital Normal University}, 
            city={Beijing},
            postcode={100048}, 
            state={Beijing},
            country={China}}

\affiliation[2]{organization={Academy for Multidisciplinary Studies},
            addressline={Capital Normal University}, 
            city={Beijing},
            postcode={100048}, 
            state={Beijing},
            country={China}}

\affiliation[3]{organization={International Information and Software Institute},
            addressline={Dalian University of Technology}, 
            city={Beijing},
            postcode={116024}, 
            state={Dalian},
            country={China}}

\begin{abstract}
Semi-supervised learning utilizes insights from unlabeled data to improve model generalization, thereby reducing reliance on large labeled datasets. Most existing studies focus on limited samples and fail to capture the overall data distribution. We contend that combining distributional information with detailed information is crucial for achieving more robust and accurate segmentation results. On the one hand, with its robust generative capabilities, diffusion models (DM) learn data distribution effectively. However, it struggles with fine detail capture, leading to generated images with misleading details. Combining DM with convolutional neural networks (CNNs) enables the former to learn data distribution while the latter corrects fine details. While capturing complete high-frequency details by CNNs requires substantial computational resources and is susceptible to local noise. On the other hand, given that both labeled and unlabeled data come from the same distribution, we believe that regions in unlabeled data similar to overall class semantics to labeled data are likely to belong to the same class, while regions with minimal similarity are less likely to. This work introduces a semi-supervised medical image segmentation framework from the distribution perspective (Diff-CL). Firstly, we propose a cross-pseudo-supervision learning mechanism between diffusion and convolution segmentation networks. Secondly, we design a high-frequency mamba module to capture boundary and detail information globally. Finally, we apply contrastive learning for label propagation from labeled to unlabeled data. Our method achieves state-of-the-art (SOTA) performance across three datasets, including left atrium, brain tumor, and NIH pancreas datasets.
\end{abstract}

\end{frontmatter}


\section{Introduction}
\label{sec:introduction}
Nowadays, medical image segmentation algorithms plays an important role in computer-assisted medical diagnosis. Image segmentation algorithms based on deep learning are state-of-the-art (SOTA) approaches and obtain good segmentation performance by training abundant labeled data. Medical images usually have characteristics of complex backgrounds and blurred boundaries, which greatly increase the difficulty of labeling, so it is not realistic to obtain a large number of labeled medical data. Semi-supervised learning attempts to make full use of both labeled and unlabeled data to improve the performance of the model.

In general, semi-supervised medical image segmentation methods include consistency  regularization (\citet{b1, b4, b2, b3, b5, b6}), pseudo-labeling (\citet{b7, b8, b9, b10, b11, b12, b13, b14, b37, b38}), adversarial learning (\citet{b15, b14}), and graph-based methods (\citet{b17, b18}). Among them, the pseudo-labeling technique leverages the model to predict pseudo-labels for unlabeled data to train the model further. \citet{b10} and \citet{b11} combined the post-processing methods of conditional random fields (CRF) and the level set to refine pseudo-labels, respectively. \citet{b9} used weak boundary annotations to assist the supervision process. \citet{b38} used contrastive learning to optimize the model. Some studies used multiple models to optimize the training process jointly. \citet{b12} defined an additional student model in a co-training framework. \citet{b14} used the mean of predictions from multiple models as pseudo-labels and introduced adversarial samples to capture differences between models. \citet{b37} suggested employing two subnets to generate pseudo-labels for a third subnet. These advanced studies have greatly improved the quality of pseudo-labels, however, a limited understanding of data and an inability to learn data distributions when using CNNs as architectures, resulting in poor generalization. 

In contrast, diffusion models is a SOTA generative model that can learn data distribution based on finite data. There are some studies for medical image segmentation utilizing diffusion models (\citet{b19, b20, b22, b21}). Some studies have employed diffusion models to generate synthetic images to expand the training dataset (\citet{b19, b20}). Other studies use diffusion models to characterize existing data. \citet{b21} used latent diffusion models to align pseudo-label distributions across different levels to improve their quality, while \citet{b22} utilized diffusion models as encoders to extract distribution-invariant features. However, diffusion models tend to focus more on learning the data distribution, often at the expense of fine details, which can result in the generation of inaccurate or misleading details.

CNNs typically focus on the details of samples. However, two main limitations arise: first, due to the restricted size of convolutional kernels, they capture only local details, making them vulnerable to local noise and lacking the ability to capture global structures. Second, capturing comprehensive high-frequency details with CNNs requires significant computational resources, which is not ideal for efficient model optimization. Mamba, conceptualized through a state-space model (SSM) (\citet{b23, b24, b25}), captures long-range dependencies and operates with linear complexity, focusing on global learning of samples to enhance the overall understanding of the model. This architecture has been increasingly used in medical image segmentation, like Segmamba \citet{b27}, VM-UNet \citet{b29}, Weak-Mamba-UNet \citet{b30}, SliceMamba \citet{b31}, and skinMamba \citet{b33}. Medical images often contain temporal or spatial relationships, making Mamba particularly well-suited for segmentation tasks that involve such structurally coherent data.

Unlabeled data, due to the lack of labels, pose a limitation for existing semi-supervised learning algorithms in effectively learning from them. Most existing methods for learning from unlabeled samples focus solely on the samples themselves, neglecting their relationship with labels. Recently, some works considered the issue \citet{b34, b6, b35, b36} and have a unified point that semantic similar regions between labeled and unlabeled data may contain the same semantics due to they come from the same distribution and their labels should be shared. Some of them used cosine similarity measurement \citet{b34} and the construction of an omni-correlation consistency matrix \citet{b6} to find similar regions between labeled and unlabeled data. \citet{b35} treated data points as nodes in a graph, with labeled and unlabeled data connected by edges. Labels propagated between nodes through edges in the graph. \citet{b36} proposed a prototype feature constraint module to constrain pixel features of unlabeled images by prototype features of labeled images, and feature alignment of the whole dataset was realized. These studies effectively provide a novel label information propagation perspective from labeled to unlabeled data. However, local semantic similarities they believed may exist across different labeled classes, making it hard to assign a class to an unlabeled region.

Inspired by these advanced works, we propose a novel semi-supervised medical image segmentation framework from the distribution perspective (Diff-CL). Firstly, Diffusion models effectively characterize the overall data distribution, but may generate images with misleading details. While integrating it with CNNs can enhance model generalization and ensure accurate detail representation. Secondly, (1) Mamba can leverage its ability to incorporate global contextual information, helping to mitigate the limitations of CNNs, which often focus heavily on local details and are prone to being affected by local noise. (2) While CNNs require substantial computational resources to capture high-frequency information, Mamba operates with linear computational complexity, making it well-suited for efficiently capturing high-frequency features. Thirdly, most existing label propagation methods focus on local semantic similarity, but such similarity can exist across different labeled classes, making it difficult to assign a class to an unlabeled region. However, in medical images, the semantic differences between classes are usually significant. Therefore, we believe that regions in unlabeled data similar to overall class semantics in labeled data are likely to belong to the same class, while regions with minimal similarity are less likely to. Our contributions include: 
\begin{itemize}
\item we propose a diffusion segmentation (DS) and convolution segmentation (CS) cross-pseudo-supervision learning mechanism and design a 3D high-frequency Mamba module to learn high-frequency details globally in medical images.
\item A new label information propagation method from labeled to unlabeled data utilizing contrastive learning.
\item Experimental results on MR and CT datasets demonstrate that Diff-CL significantly improves model generalization.  
\end{itemize}

The paper is organized as follows. Section \ref{section 2} reviews related works. Section \ref{section 3} gives preliminaries of Diff-CL needed. Section \ref{section 4} introduces our method in detail. Section \ref{section 5} presents experiment results on four medical image datasets. Section \ref{section 6} is a summary of our whole research. 

\begin{figure*}[!t]
\center
\includegraphics[scale=.26]{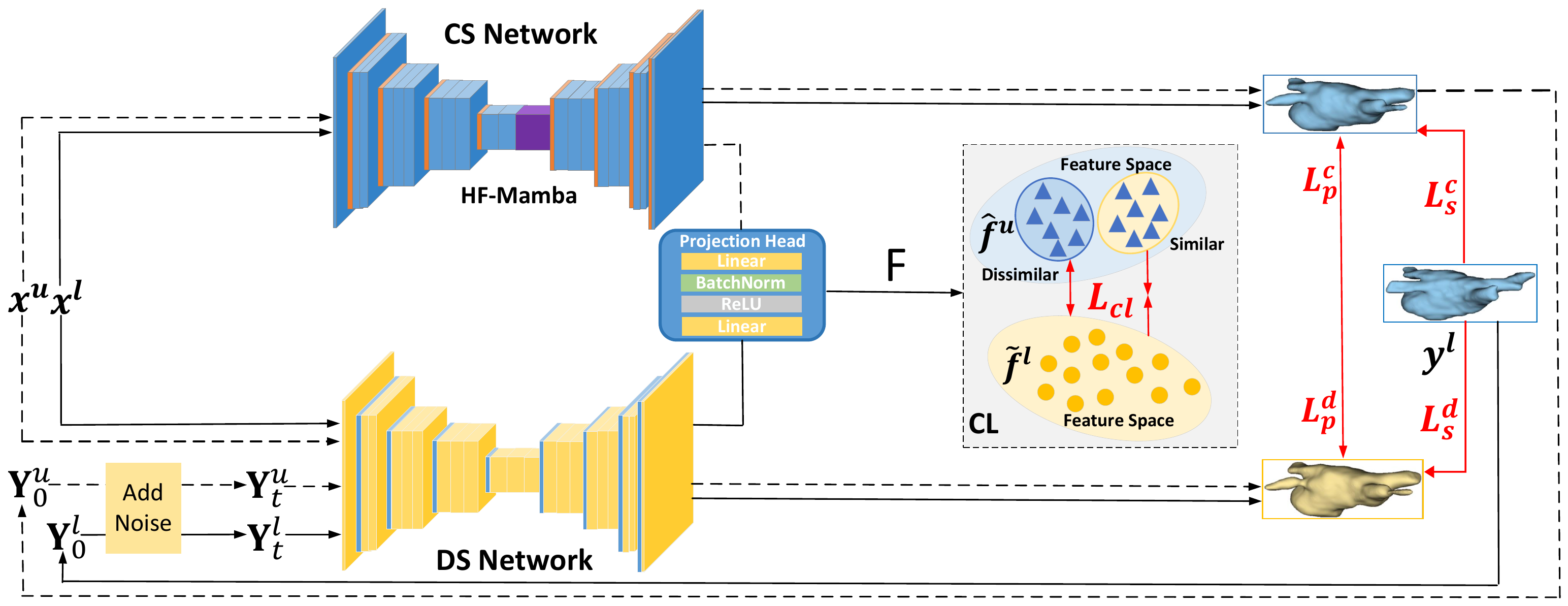}
\caption{The overall overview of Diff-CL. The backbone consists of CS and DS networks with a projection head. HF-Mamba is our high-frequency mamba module. $x^l$ and $x^u$ are labeled and unlabeled data. $y^l$ is the label of labeled data. $Y^l_0 = y^l$ and $Y^u_0$ is the pseudo-label of $x^u$ from CS network. After adding t-step Gaussian noise, we get $Y^l_t$ and $Y^u_t$. 
$(x^l, Y^l_t)$ and $(x^u, Y^u_t)$ are concatenated as inputs of DS network. $(\tilde{f}^l, \hat{f}^u)$ are labeled and unlabeled features from DS and CS networks respectively. Inter-CL represents inter-sample contrastive learning which closes similar features and pushes dissimilar features away between $\tilde{f}^l$ and $\hat{f}^u$. There are supervised and unsupervised losses. The supervised loss includes $L^c_s$ and $L^d_s$. The unsupervised loss includes cross pseudo-supervision losses $L^c_p$ and $L^d_p$ and inter-sample contrastive learning loss $L_{cl}$.}
\label{fig1}
\end{figure*}

\section{Related Works}
\label{section 2} 

We introduce the diffusion model and contrastive learning methods for semi-supervised medical image segmentation and mamba-based methods for medical image segmentation.

\subsection{Diffusion Model Methods}
\label{section 2.1}

\citet{b19} used a latent diffusion model to generate synthetic medical images, reducing the burden of data annotation and addressing privacy issues associated with medical data collection. \citet{b21} introduced a latent diffusion label correction model (DiffRect) for semi-supervised medical image segmentation. A Latent Feature Rectification (LFR) module is applied in the latent space, leveraging latent diffusion to align pseudo-label distributions at different levels. It uses a denoising network to learn continuous distribution transitions from coarse to fine. \citet{b20} employed an implicit diffusion model (LDM) to generate new scleral images. \citet{b22} proposed an aggregation and decoupling framework, where the aggregation part constructs a common knowledge set using a diffusion encoder to extract distribution-invariant features from multiple domains. The decoupling part, consisting of three decoders, separates the training process for labeled and unlabeled data, thus preventing overfitting to labeled data, specific domains, or classes.

Different from them, we propose a dual DS-CS cross pseudo-supervision learning mechanism. The DS network learns the data distribution while the CS network provides detailed calibration. When learning data distribution, DM tends to reconstruct the global structure and main features of the data rather than the details, so the recovery of details may be insufficient. CNNs focus on details of samples. Therefore, the architecture of combining DM and CNNs complements each other.

\subsection{Mamba-based Methods}
\label{section 2.2}

Many studies have explored applications of Mamba in medical image segmentation. Weak-Mamba-UNet \citet{b30} combined CNN-based UNet \citet{b42} for detailed local feature extraction, a Swin Transformer-based SwinUNet \citet{b43} for comprehensive global context understanding, and a VMamba-based Mamba-UNet \citet{b41} for efficient long-range dependency modeling. SliceMamba \citet{b31} introduced a Bidirectional Slice Scanning (BSS) module with a specific mechanism, enhancing the ability of Mamba to model local features. SkinMamba \citet{b33} introduced the Frequency Boundary Guided Module (FBGM) to achieve high-frequency restoration and boundary prior guidance. However, these methods are not specifically designed for 3D medical image segmentation. SegMamba \citet{b27} introduced a Three-directional Mamba (ToM) module to enhance sequential modeling of 3D features from three different directions. 

Unlike them, we develop a 3D high-frequency Mamba module for capturing high-frequency details by leveraging global contextual information. Medical images often exhibit structured anatomical patterns with a clear sequential relationship, where global context plays a key role in improving the extraction of fine details.

\subsection{Contrastive Learning methods}
\label{section 2.3}

Contrastive learning is to learn distinctive feature representations by increasing the margin between different classes in the feature space, thus facilitating the distinction between positive and negative pairs. Pixel contrastive learning has been widely adopted for semi-supervised segmentation tasks (\citet{b44, b45, b46, b38, b48}). \citet{b44} introduced pixel contrastive learning with confidence sampling into consistency training for semi-supervised segmentation. \citet{b46} proposed generating two different views from the same input, encouraging consistency in overlapping regions while establishing bidirectional contrast in other regions. 
\citet{b38} introduced a bidirectional voxel contrastive learning strategy to tackle insufficient class separability. This strategy further optimized the model to improve class separability by pulling voxels of the same class closer in feature space while pushing apart those of different classes. \citet{b48} designed a Pseudo-label Guided Contrastive Loss (PLGCL) to leverage contrastive learning in capturing important class-discriminative features.

In contrast, we use contrastive learning to propagate label information from labeled to unlabeled data by narrowing the boundaries between regions in the unlabeled data similar to the class overall semantics of labeled data, and expanding the boundaries between unrelated regions. This enhances the intra-class similarity and inter-class separability of the unlabeled data.

\section{Preliminaries}
\label{section 3}
For completeness and readability, we introduce the knowledge of diffusion model and Mamba in this section.

\subsection{Denoising Diffusion Implicit Models (DDIM)}
\label{section 3.1}

As a variant of diffusion models, DDIM \citet{b55} includes a forward noise-adding process and an accelerated reverse denoising process. The forward process of diffusion models is to define a forward Markov chain of Gaussian transitions, which gradually adds noise to the data \( x_0 \), defined as follows:
\begin{align}
x_t = \sqrt{\bar{\alpha}_t}\cdot{x_0} + \sqrt{1-\bar{\alpha}_t}\cdot{\epsilon}, \quad \epsilon \in{\mathcal{N}(0,1)},
\end{align}
where $ x_0 $ is the original data, and $ t $ is the diffusion timestep. $\bar{\alpha}_t = \Pi_{i=1}^t{\alpha_i}$ and $ \alpha_t $ is a predefined time-dependent scalar.

In DDIM, the reverse denoising process significantly reduces the number of required steps while still achieving high-quality sample generation. The reverse process of DDIM :
\begin{align}
x_{\tau_i-1}=\sqrt{\alpha_{\tau_i-1}}x_0+\sqrt{1-\alpha_{\tau_i}}\cdot{\frac{x_{\tau_i}-\sqrt{\alpha_{\tau_i}}x_0}{\sqrt{1-\alpha_{\tau_i}}}}, \forall i \in{[S]},
\end{align}
where $\tau_i$ is a dub-sequence of $[1,...,T]$ of length $S$ with $\tau_S=T$. Compared to traditional diffusion models that predict noise \( \epsilon_\theta \), DDIM allows the network to predict the image \( x_0 \) directly, given by:

\begin{equation}
x_0 = \frac{x_t - \sqrt{1 - \bar{\alpha}_t} \epsilon_\theta(x_t, t)}{\sqrt{\bar{\alpha}_t}}.
\end{equation}
This approach enables the model to generate samples deterministically without conducting iterative noise prediction. 

\begin{figure*}[!t]
\center
\includegraphics[scale=.37]{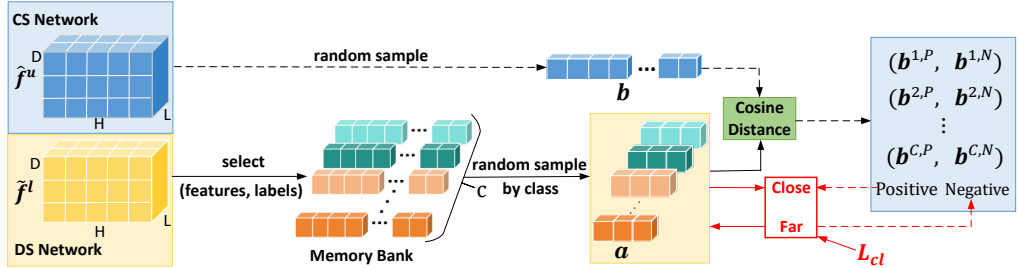} 
\caption{$\tilde{f}^l$ and $\hat{f}^u$ are labeled and unlabeled features output by DS and CS networks respectively. $C$ is the class number of a dataset. $\textbf{a}$ and $\textbf{b}$ are randomly sampled feature vectors from our memory bank and $\hat{f}^u$, respectively. By calculating cosine distances between $\textbf{a}$ and $\textbf{b}$, we get unlabeled positive and negative pairs $\textbf{b}^{\gamma, P}$ and $\textbf{b}^{\gamma, N}$ of $\textbf{a}$,where $\gamma \in{(1,...,C)}$. $L_{cl}$ is the inter-sample contrastive loss.}
\label{fig2}
\end{figure*}

\subsection{Mamba}
\label{section 3.2}
For the convenience of readers, we present the definition of Mamba. 
Mamba \citet{b24} is built on the selective state space model (SSM). It is characterized by parameters, $\bar{A}$, $\bar{B}$, and $C$, and transforms a given input sequence $(x_t)_{t=1}^T$ into an output sequence of the same size $(y_t)_{t=1}^T$ via the following equations:
\begin{equation}
\begin{aligned}
    & h_t = \bar{A}_t h_{t-1} + \bar{B}_t x_t  \\
    & y_t = C_t h_t 
\end{aligned}
\end{equation}
where the initial state $h_0 = 0$. What distinguishes the selective SSM from the original SSM (S4) \citet{b26} is that the evolution parameter, $\bar{A}_t$, and projection parameters, $\bar{B}_t$ and $C_t$, are functions of the input $x_t$. This enables dynamic adaptation of the SSM’s parameters based on input. This mechanism dynamicity facilitates the model focusing on relevant information while ignoring irrelevant details when processing a sequence.

\section{Methodology}
\label{section 4}
Our overall framework is shown in Fig. \ref{fig1}. The proposed Diff-CL has dual DS-CS networks as our backbone that includes a diffusion segmentation network $\tilde{F}(\cdot)$ and a convolutional segmentation network $\hat{F}(\cdot)$. A projection head is used after DS and CS networks to derive labeled features $\tilde{f}^l$ and unlabeled features $\hat{f}^u$, respectively. The dimension of features is $d$. In this section, we first present our dual DS-CS cross-pseudo-supervision mechanism in section \ref{section 4.1}, and the high-frequency Mamba module in section \ref{section 4.2}. Then our label propagation method by contrastive learning is explained in section \ref{section 4.3}. Finally, semi-supervised training is detailed in section \ref{section 4.4}. 

\begin{table*}[!t]
\centering
\caption{Notations. (the same hereinafter)}
\renewcommand{\arraystretch}{1.2}
\setlength{\tabcolsep}{4.2pt}
\begin{tabular}{|p{6.4cm}|p{6.5cm}|}
\hline
\multicolumn{2}{|c|}{Dataset} \\
\hline
$S=S^l\cup{S^u}$ & Dataset \\ $S^l=\left\{x^l_{i}, y^l_{i}\right\}_{i=1}^M$ & Labeled dataset \\
$S^u=\left\{x^u_{j}\right\}_{j=1}^N$ & Unlabeled dataset  \\ 
$\left\{x^l_{i}, x^u_{j}, y^l_{i}\right\}\in{R^{H\times{L}\times{D}}}$ & Dimension of data \\
$C$ & Class number of one dataset \\
\hline
\multicolumn{2}{|c|}{Network} \\
\hline
$\tilde{F}(\cdot)$ & DS network \\ $\hat{F}(\cdot)$ & CS network \\ 
$F$ & Projection head \\  $\tilde{f}^l$ & Labeled features output by our projection head \\
$\hat{f}^u$ & Unlabeled features output by our projection head \\
$\hat{Y}^u$ & Pseudo-labels of unlabeled samples predicted by CS network \\ 
$\tilde{Y}^u$ & Pseudo-labels of unlabeled samples predicted by DS network \\
\hline
\multicolumn{2}{|c|}{Label propagation by contrastive learning}
\\
\hline
$\textbf{a}=(\textbf{a}^1, \textbf{a}^2, ..., \textbf{a}^C)$, where $\textbf{a}^{\gamma}= (a^{\gamma}_1, a^{\gamma}_2,... a^{\gamma}_p)$, $\gamma \in{\left\{\ 1,..., C \right\}}$ & Randomly sampled labeled feature vector from DS \\ 
$\textbf{b}=(b_1, b_2, ..., b_{q})$ & Randomly sampled unlabeled feature vector from CS \\
\hline
\end{tabular}
\label{table6}
\end{table*}

\subsection{DS-CS Cross-Pseudo-Supervision Learning}
\label{section 4.1}

\textbf{DS-CS} With the specially designed, cross-pseudo-supervised learning means that pseudo-labels of DS provide CS distribution information, and pseudo-labels of CS provide DS correct details.

\textbf{CS} A mini-batch of data inputs in one forward pass including $m$ labeled samples $\left\{x^l_i\right\}_{i=1}^m$ and $n$ unlabeled samples $\left\{x^u_j\right\}_{j=1}^n$. Softmax outputs of CS for them are denoted as $\hat{P}^l=\left\{\hat{P}^l_i \right\}_{i=1}^m$ and $\hat{P}^u=\left\{\hat{P}^u_j \right\}_{j=1}^n$. Pseudo-labels of the softmax outputs are denoted as $\hat{Y}^l=\left\{\hat{Y}^l_i \right\}_{i=1}^m$ and $\hat{Y}^u=\left\{\hat{Y}^u_j \right\}_{j=1}^n$.

\textbf{DS} DS diffuses labels with samples as the condition. However, there is no label for unlabeled data. We replace it with pseudo-labels $\hat{Y}^u$ obtained from CS. To get inputs of DS, first, $y^l=\left\{y^l_i \right\}_{i=1}^m$ and $\hat{Y}^u$ are converted into one-hot forms, denoted as $Y^l_{0}$ and $Y^u_{0}$, which serve as the initial clean labels of DS. Next, successive t-step noise $\epsilon$ are added to them to obtain noise labels:
\begin{align}
Y^{l/u}_t = \sqrt{\bar{\alpha}_t}Y^{l/u}_0 + \sqrt{1-\bar{\alpha}_t}\epsilon, \epsilon \in{\mathcal{N}(0,1)},
\label{add noise}
\end{align}
where $\bar{\alpha}_t$ is described in \ref{section 3.1}.
Finally, $(x^l, Y^l_t)$ and $(x^u, Y^u_t)$ are concatenated as inputs of DS. Softmax outputs of DS for them are denoted as $\tilde{P}^l=\left\{\tilde{P}^l_i \right\}_{i=1}^m$ and $\tilde{P}^u=\left\{\tilde{P}^u_j \right\}_{j=1}^n$. Pseudo-labels of the softmax outputs are denoted as $\tilde{Y}^l=\left\{\tilde{Y}^l_i \right\}_{i=1}^m$ and $\tilde{Y}^u=\left\{\tilde{Y}^u_j \right\}_{j=1}^n$.

\textbf{Cross pseudo-supervision.} Supervise softmax predictions of DS with pseudo-labels of CS $\hat{Y}^u$ and supervise softmax predictions of CS with pseudo-labels of DS $\tilde{Y}^u$:
\begin{align}
L^d_p=D(\tilde{P}^u, \hat{Y}^u)+\lambda_1{E(\tilde{P}^u, \hat{Y}^u)},
\label{DM cross}
\end{align}
\begin{align}
L^c_p=D(\hat{P}^u, \tilde{Y}^u)+\lambda_2{E(\hat{P}^u, \tilde{Y}^u)},
\label{CNN cross}
\end{align}
where $\lambda_1$ and $\lambda_2$ are weights of $L^d_p$ and $L^c_p$ and
\begin{align}
D(\tilde{P}^u, \hat{Y}^u)=1-\frac{2|\tilde{P}^u \cap{\hat{Y}^u}|}{|\tilde{P}^u|+|\hat{Y}^u|},
\label{Dice loss}
\end{align}
\begin{align}
E(\tilde{P}^u, \hat{Y}^u)=-\sum_{i=0}^{c-1}{\hat{Y}^u_i}\cdot{log(\tilde{P}^u_i}),
\label{CE loss}
\end{align}
where $D(\hat{P}^u, \tilde{Y}^u)$ and $E(\hat{P}^u, \tilde{Y}^u)$ are the same with them.

\subsection{High-Frequency Mamba Module (HFM)}
\label{section 4.2}
Medical images often display structured anatomical patterns with clear sequential relationships, where global contextual information is crucial for enhancing the extraction of fine details. To address the computational challenges faced by CNNs in capturing high-frequency details, we develop a 3D high-frequency Mamba module that leverages the global contextual information of the images to learn details of samples globally.

\begin{figure}[!t]
\center
\includegraphics[scale=.42]{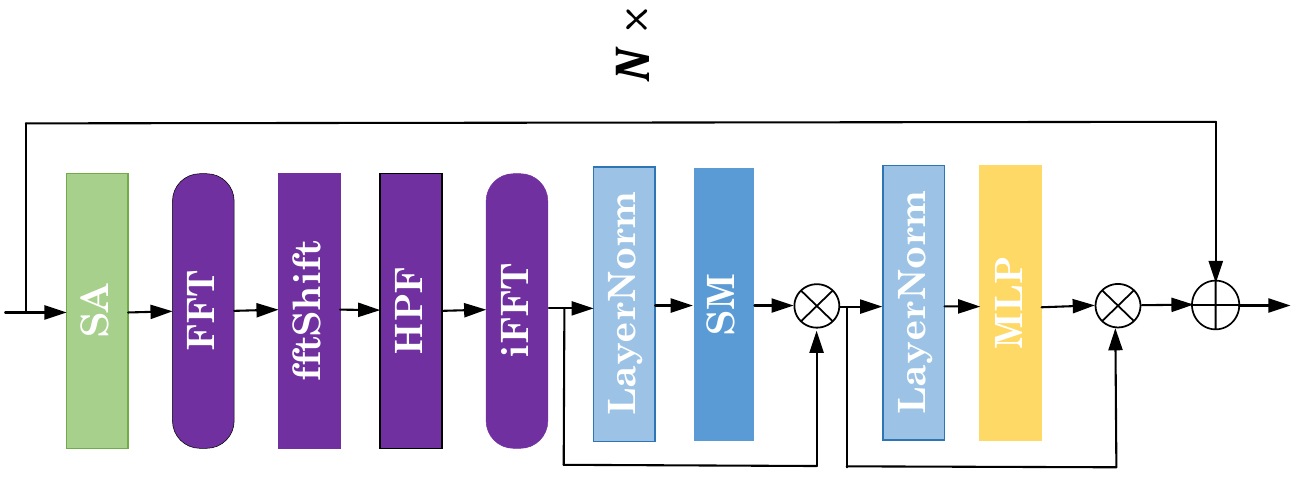} 
\caption{High-frequency mamba block.}
\label{fig3}
\end{figure}

To obtain high-frequency components of features, 3D features are first transformed from the spatial domain to the frequency domain by the fast Fourier transform (FFT). Then, the features are filtered through high-pass filtering (HPF) to obtain high-frequency features. Finally, the obtained high-frequency features are transformed into the spatial domain for mamba attention learning. The process flow is shown in Figure \ref{fig3}.

The module is defined as follows:
\begin{equation}
\begin{aligned}
& F_g = \mathrm{FFT}(\mathrm{SA}(\hat{f}^{l/u})) \\
& F_h = \mathrm{iFFT}(\mathrm{HPF}(\mathrm{fftShift}((F_g)))) \\
& F_m = \mathrm{SM}(\mathrm{LN}(F_h))+F_h \\
& F_o = \mathrm{MLP}(\mathrm{LN}(F_m))+F_m,
\end{aligned}
\end{equation}
where $\hat{f}^{l/u}$ denotes 3D labeled or unlabeled features of CS network. We adopt a Spatial Attention module ($\mathrm{SA}$) and a Space Mamba ($\mathrm{SM}$) introduced in \citet{b27}. fftShift represents spectrum centralization. iFFT is Fast Inverse Fourier transform. $\mathrm{SM}$ includes forward, backward, and space orientation for encoding 3D features. LN and MLP denote the Layer Normalization and Multiple Layers Perception Layer to enrich the feature representation.

\subsection{Label Information Propagation by Contrastive Learning}
\label{section 4.3}

Medical images typically exhibit consistent structures within the same class and significant differences between different classes. We believe that regions in unlabeled data similar to overall class semantics in labeled data are likely to belong to the same class, while regions with minimal similarity are less likely to. We uses contrastive learning to label information propagation by narrowing the boundaries between regions in unlabeled data similar to overall semantics of labeled data in each class and expand those between unrelated ones.

A memory bank (MB) is built to store pairs of labeled features from the DS and corresponding labels. Screen out the part that DS predicts correctly for labeled samples, and store the features of these pixels in MB. We randomly select $p$ labeled features for each class from MB which represent class semantics and get $\textbf{a}=(\textbf{a}^1, \textbf{a}^2, ..., \textbf{a}^C)$, where $\textbf{a}^{\gamma}= (a^{\gamma}_1, a^{\gamma}_2,... a^{\gamma}_p)$, $\gamma \in{\left\{\ 1,..., C \right\}}$, where $C$ is the class number of a dataset. We randomly sample $q$ unlabeled features $\textbf{b}=(b_1,b_2,...,b_{q})$ from all unlabeled features projected by CS.

Next, find out the unlabeled features that are semantically similar and unrelated to each class in labeled features from the unlabeled features $\textbf{b}$. Calculate the cosine distance between each unlabeled feature in $\textbf{b}$ and each labeled feature in $\textbf{a}^{\gamma}$:
\begin{align}
    cos(b_j, a_i) = \frac{\sum_{w=1}^r{a_i\cdot{b_j}}}{\sum_{k=1}^r{a_i}\times{\sum_{k=1}^r{b_j}}},
\end{align}
where $a_i \in{\textbf{a}^{\gamma}}$ and $b_i\in{\textbf{b}}$. $r$ is the dimension of the feature. 
According to cosine similarity, find the top-k most similar and top-k least similar unlabeled features in $\textbf{b}$ to the labeled feature vector $\textbf{a}$:
\begin{align}
\textbf{b}^{\gamma,P} = \mathop{max}\limits_{j=1}^k{\sum_{i=1, a_i\in{\textbf{a}^{\gamma}}}^{p}{cos(b_j, a_i)}}, 
\end{align}
\begin{align}
\textbf{b}^{\gamma,N} = \mathop{max}\limits_{j=1}^k{\sum_{a_i\in{\textbf{a}^{\gamma}}}^{p}{(1-cos(b_j, a_i))}}, 
\end{align}
Every feature in $\textbf{b}^{\gamma,P}$ and $\textbf{b}^{\gamma,N}$ are positive and negative pairs of labeled feature vector $\textbf{a}^{\gamma}$, respectively. We can get $\left\{(\textbf{b}^{1,P}, \textbf{b}^{1,N}), ..., (\textbf{b}^{C,P}, \textbf{b}^{C,N}) \right\}$.

Contrastive learning narrows the distance between positive pairs and enlarges the distance between negative pairs. For $a_i \in \textbf{a}^{\gamma}$ and $b_j \in \textbf{b}^{\gamma, N}$, the contrastive loss between them is calculated as follows:
\begin{align}
  \mathcal{L}(a_i,b_j) = -\log \frac{e^{\cos(a_i, b_j)/\tau}}{e^{\cos(a_i, b_j)/\tau} + \sum_{\alpha=1, b_{\alpha} \in \textbf{b}^{\gamma, N}}^{k} e^{\cos(a_i, b_{\alpha})/\tau}},  
\label{contrastive loss}
\end{align}
where $a_i \in \textbf{a}^{\gamma}$ and $b_j \in \textbf{b}^{\gamma, N}$. $\tau$ is a temperature parameter. Our label information propagation loss is defined as follows:
\begin{align}
  L_{cl} = \sum_{\gamma=1}^{C}{\sum_{\left\{i=1, i\in{\textbf{a}^{\gamma}}\right\}}^{p}{\sum_{\left\{j=1, j\in{\textbf{b}^{\gamma, P}}\right\}}^{k}{\mathcal{L}(a_i, b_j)}}}.
\label{Inter-sample contrastive loss}
\end{align}

\subsection{Semi-supervised Training}
\label{section 4.4}

\begin{algorithm}[!t]
\renewcommand{\algorithmicrequire}{\textbf{Input:}}
\renewcommand{\algorithmicensure}{\textbf{Output:}}
\caption{\textbf{Training Process of Diff-CL.}} \label{algorithm1}
\begin{algorithmic}[1] 
\REQUIRE CS $f^{c}(\cdot)$; DS $f^{d}(\cdot)$; Labeled dataset $S^l$; Unlabeled dataset $S^u$; the max iteration $I$
\ENSURE Trained CS $f^c(\cdot)$ for inference
\STATE Initialize CS $f^c(\cdot)$ and DS $f^d(\cdot)$;
\FOR {$epoch \leq I$}
\STATE Sampled batch data $x^l$ and $x^u$ from $S^l$ and $S^u$; 
\STATE CS Forward: $\hat{P}^{l} = f^c(x^l)$; $\hat{P}^{u} = f^c(x^u)$;
\STATE Pseudo labels of CS:\\ $\hat{Y}^l= argmax(\hat{P}^{l})$; $\hat{Y}^u= argmax(\hat{P}^{u})$;
\STATE Add Noise by Eq. \ref{add noise}: $y^l_t$; $y^u_t$;
\STATE DS Forward: $\tilde{P}^{l} = f^d(x^l, y^l_t)$; $\tilde{P}^{u} = f^d(x^u, y^u_t)$;
\STATE Get supervised loss of CS by Eq. \ref{CS sup};
\STATE Get supervised loss of DS by Eq. \ref{DS sup};
\STATE Cross pseudo-supervision loss of CS by Eq. \ref{CNN cross};
\STATE Cross pseudo-supervision loss of DS by Eq. \ref{DM cross};
\STATE Label information propagation loss by Eq. \ref{Inter-sample contrastive loss};
\STATE Calculate loss $L^c_u$ by Eq. \ref{CNN unsup};
\STATE Update CS using Eq. \ref{CNN loss};
\STATE Update DS using Eq. \ref{DM loss};
\ENDFOR
\end{algorithmic}
\label{alg1}
\end{algorithm}

The total objective function to train our DS-CS networks is both a weighted combination of a supervised loss on labeled data and an unsupervised loss on unlabeled data, defined as: 
\begin{align}
L^{d}=L^d_{s}+\mu_1{L^d_{p}}, \label{DM loss}\\
L^{c}=L^c_{s}+\mu_2{L^c_{u}}, \label{CNN loss}
\end{align}
where $L^{d}$ and $L^{c}$ are the total losses of DS and CS, respectively. $\mu_1$ and $\mu_2$ are the weights to adjust ratios of cross pseudo-supervision loss for DS $L^d_{p}$ and unsupervised loss for CS $L^c_{u}$, respectively. 

The dice coefficient $D(\cdot)$ and cross-entropy $E(\cdot)$ guide the training for labeled data:
\begin{align}
L^d_s=D(\tilde{P}^l, y^l)+\beta_1{E(\tilde{P}^l, y^l)},
\label{DS sup}
\end{align}
\begin{align}
L^c_s=D(\hat{P}^l, y^l)+\beta_2{E(\hat{P}^l, y^l)},
\label{CS sup}
\end{align}
where $\beta_1$ and $\beta_2$ are the weights of cross entropy $E(\tilde{P}^l,y^l)$ and $E(\hat{P}^l,y^l)$ and
\begin{align}
D(\tilde{P}^l, y^l)=1-\frac{2|\tilde{P}^l \cap{y^l}|}{|\tilde{P}^l|+|y^l|},
\label{DM sup}
\end{align}
\begin{align}
E(\tilde{P}^l, y^l)=-\sum_{i=0}^{c-1}{y^l_i}\cdot{log(\tilde{P}^l_i}),
\end{align}
where $D(\hat{P}^l, y^l)$ and $E(\hat{P}^l, y^l)$ are the same with them.

Cross pseudo supervision loss $L^d_p$ and $L^c_p$ in section \ref{section 4.1} and inter-sample contrastive learning loss $L_{cl}$ in section \ref{section 4.3} guide the DS and CS training for unlabeled data, defined as:
\begin{align}
L^c_u=L^c_{p} + \eta{L_{cl}},
\label{CNN unsup}
\end{align}
where $\eta$ are weights of $L_{cl}$. The entire flow of Diff-CL is presented in Algorithm \ref{algorithm1}.

\section{Experiments}
\label{section 5}

\begin{table}[!t]
\centering
\caption{Parameter Settings of Diff-CL on three datasets. $\mu_1$, $\mu_2$, $\beta_1$, $\beta_2$, $\lambda_1$, $\lambda_2$, and $\eta$ are introduced in \ref{section 4}. HF is the threshold of high-frequency filters.}
\renewcommand{\arraystretch}{1.2}
\setlength{\tabcolsep}{5.2pt}
\scalebox{0.75}{
\begin{tabular}{@{}lcccccccccc@{}}
\hline
Dataset & Labeled/Unlabeled & Batch Size & $\mu_1$ & $\mu_2$ & $\beta_1$ & $\beta_2$ & $\lambda_1$ & $\lambda_2$ & $\eta$ & HF \\
\hline
\multirow{2}{*}{Pancreas} & 6/56 & 4 & 1 & 1 & $0.1\times{\lambda{(t)}}$ & $0.1\times{\lambda{(t)}}$ & 1 & 1 & 0.007 & 0.8 \\
& 12/50 & 4 & 1 & 1 & $0.1\times{\lambda{(t)}}$ & $0.1\times{\lambda{(t)}}$ & 1 & 1 & 0.0003 & 0.5 \\
\multirow{2}{*}{LA} & 4/76 & 4 & 1 & 1 & $0.01\times{\lambda{(t)}}$ & $0.01\times{\lambda{(t)}}$ & 1 & 1 & 0.7 & 0.6 \\ 
 & 8/72 & 4 & 1 & 1 & $0.01\times{\lambda{(t)}}$ & $0.01\times{\lambda{(t)}}$ & 1 & 1 & 0.5 & 0.7 \\
\multirow{2}{*}{BraTS} & 25/225 & 4 & 1 & 1 & $0.1\times{\lambda{(t)}}$ & $0.1\times{\lambda{(t)}}$ & 1 & 1 & 0.007 & 0.5  \\
& 50/250 & 4 & 1 & 1 & $0.1\times{\lambda{(t)}}$ & $0.1\times{\lambda{(t)}}$ & 1 & 1 & 0.003 & 0.5  \\
\hline
\end{tabular}}
\label{table1}
\end{table}

\subsection{Datasets and experimental setup}
We evaluate the performances of Diff-CL on three publicly available datasets.

\textbf{Left atrium segmentation dataset.}
3D Left Atrium (LA) segmentation
dataset \citet{b50} contains 100 3D gadolinium-enhanced
MR Images with a resolution of $0.625mm\times{0.625mm}\times{0.625mm}$. The preprocessing procedure is the same as in (\citet{b1, b2, b3, b5, b6, b7}) and we use 80 scans for training and 20 scans for testing. All images are cropped at the center of the heart region and normalized to zero mean and unit variance.  

\textbf{Brain tumor segmentation dataset:}
The brain tumor segmentation (BraTS) is from the BraTS 2019 challenge \citet{b51}, containing preoperative MRI (with modalities of T1, T1Gd, T2, and T2-FLAIR) of 335 glioma patients,
where 259 patients are with high-grade glioma (HGG) and 76 patients
are with low-grade glioma (LGG). Like \citet{b5}, we use T2-FLAIR images in this task and segment the whole tumor region. The images are resampled to the isotropic resolution of 1 × 1 × 1 mm3. Following the same data split and preprocessing procedure as in \citet{b5}, 250 samples are used for training, 25 for validation, and the remaining 60 for testing. 

\begin{table}[!t]
\centering
\caption{Results of quantitative comparison on LA dataset.}
\renewcommand{\arraystretch}{1.2}
\setlength{\tabcolsep}{1mm}
\setlength{\tabcolsep}{5.2pt}
\begin{threeparttable} 
\scalebox{0.78}{
\begin{tabular}{@{}lccccc@{}}
		\hline
		\multirow{2}{*}{Method} & Scans used & \multicolumn{4}{c}{Metrics}\\
		\cline{2-6}
		     & Labeled/Unlabeled & Dice($\%$) $\uparrow$ & Jaccard($\%$) $\uparrow$ & 95HD $\downarrow$ & ASD $\downarrow$ \\
		\hline
            V-Net \citet{b54} & 4/0 & 52.55 & 39.60 & 47.05 & 9.87 \\
            \hline
		DTC \citet{b1} & 4/76 & 82.75 & 71.55 & 13.77 & 3.91	\\ 
		SASSnet \citet{b2} & 4/76 & 83.26 & 71.92 & 15.51 & 4.63 \\
		UA-MT \citet{b3} & 4/76 & 81.16 & 68.97 & 24.22 & 6.97 \\
		URPC \citet{b4} & 4/76 & 83.47 & 72.56 & 14.02 & 3.68 \\
		MC-Net \citet{b7} & 4/76 & 84.06 & 73.04 & 12.16 & 2.42 \\
		MC-Net+ \citet{b8} & 4/76 & 83.13 & 71.58 & 12.69 & 2.71 \\
		AC-MT \citet{b5} & 4/76 & 87.42 & 77.83 & \underline{9.09} & \underline{2.19} \\
		CAML \citet{b6} & 4/76 & \underline{87.54} & \underline{77.95} & 10.76 & 2.58 \\
ML-RPL \citet{b53} & 4/76 & 84.70 & 73.75 & 13.73 & 3.53 \\
Diff-CL (ours) & 4/76 & \textbf{89.03} & \textbf{80.35} & \textbf{6.36} & \textbf{2.18} \\
\hline
V-Net \citet{b54} & 8/0 & 78.57 & 66.96 & 21.20 & 6.07 \\
\hline
DTC \citet{b1} & 8/72 & 87.43 & 78.06 & 8.38 & 2.40	\\ 
SASSnet \citet{b2} & 8/72 & 87.80 & 76.91 & 14.57 & 4.11 \\
UA-MT \citet{b3} & 8/72 & 84.48 & 73.98 & 17.13 & 4.82	\\
URPC \citet{b4} & 8/72  & 87.14 & 77.41 & 11.79 & 2.43	\\
MC-Net \citet{b7} & 8/72 & 88.99 & 80.32 & \underline{7.92} & \underline{1.76} \\
MC-Net+ \citet{b8} & 8/72 & 88.33 & 79.32 & 9.07 & 1.82 \\
AC-MT \citet{b5} & 8/72 & 88.74 & 79.94 & 8.29 & 1.91 \\
CAML \citet{b6} & 8/72 & \underline{89.44} & \underline{81.01}  & 10.10 & 2.09 \\
ML-RPL \citet{b53} & 8/72 & 87.35 & 77.72 & 8.99 & 2.17 \\
Diff-CL (ours) & 8/72 & \textbf{91.00} & \textbf{83.54} & \textbf{5.08} & \textbf{1.68} \\
\hline
V-Net \citet{b54} & 100$\%$ & 91.62 & 84.60 & 5.40 & 1.64 \\
\hline
\end{tabular}}
\label{table2}
\end{threeparttable} 
\end{table}

\textbf{NIH pancreas dataset}
The NIH Pancreas Dataset (Pancreas) \citet{b49} includes 82 contrast$-$enhanced abdominal 3D CT volumes with manual annotation. The size of each CT volume range from $512\times{512}\times{181}$ to $512\times{512}\times{466}$. Our preprocessing is like \citet{b1}. In our experiments, the soft tissue CT window we use is $[{-}120, 240]$ HU, and the CT scans are cropped centered at the pancreas region, and the margins are enlarged with 25 voxels. 62 volumes are used for training and 20 volumes are used for testing.

\textbf{Metrics.}
We use four complementary evaluation metrics to quantitatively evaluate the segmentation performances of every model during the testing process, including the Dice similarity Index (Dice), Jaccard Index (Jaccard), Average surface distance (ASD), and 95$\%$ Hausdorff (95HD). Dice and Jaccard measure region matching, and higher Dice and Jaccard scores indicate higher pixel-level matching. ASD and 95HD are two boundary-based metrics that measure boundary differences in images, and lower ASD and 95HD scores indicate more minor boundary differences in images.

\textbf{Baseline approaches.}
We compare Diff-CL with seven semi-supervised image segmentation methods: dual-task consistency (DTC) \citet{b1}, shape-aware model (SASSNet) \citet{b2}, uncertainty-aware MT (UA-MT) \citet{b3}, uncertainty rectified pyramid consistency (URPC) \citet{b4}, Mutual Consistency Network (MC-Net) \citet{b7}, MC-Net+ \citet{b8}, ambiguity-consensus mean-teacher (AC-MT) \citet{b5}, correlation-aware mutual learning (CAML) \citet{b6}, and mutual learning with reliable pseudo label (ML-RPL) \citet{b53}. 

\begin{table}[!t]
\centering
\caption{Results in quantitative comparison on BraTS dataset.}
\renewcommand{\arraystretch}{1.3}
\setlength{\tabcolsep}{5.2pt}
\scalebox{0.80}{
\begin{tabular}{@{}lccccc@{}}
\hline
\multirow{2}{*}{Method} & Scans used & \multicolumn{4}{c}{Metrics}\\
\cline{2-6}
 & Labeled/Unlabeled & Dice($\%$)$\uparrow$ & Jaccard($\%$)$\uparrow$ & 95HD$\downarrow$ & ASD$\downarrow$ \\
\hline
V-Net \citet{b54} & 25/0 & 74.43 & 61.86 & 37.11 & 2.79 \\
\hline
DTC \citet{b1} & 25/225 & 80.01 & 69.78 & 11.56 & 1.94 \\ 
SASSnet \citet{b2} & 25/225 & 77.44 & 66.24 & 21.55 & 6.77 \\
UA-MT \citet{b3} & 25/225 & 82.82 & 72.42 & 14.01 & 3.75 \\
URPC \citet{b4} & 25/225 & \underline{83.61} & \underline{73.52} & \underline{9.92} & \underline{1.59} \\
MC-Net \citet{b7} & 25/225 & 83.33 & 73.29 & \underline{9.92} & 2.42 \\
MC-Net+ \citet{b8} & 25/225 & 82.18 & 72.18 & 10.81 & 1.74\\
AC-MT \citet{b5} & 25/225 & 81.60 & 71.04 & 10.76 & 2.06 \\
CAML \citet{b6} & 25/225 & 83.91 & 74.18 & 13.23 & 3.54 \\
ML-RPL \citet{b53} & 25/225 & 81.60 & 71.23 & 11.63 & 2.99 \\
Diff-CL (ours) & 25/225 & \textbf{84.63} & \textbf{75.05} & \textbf{12.08} & \textbf{3.73} \\
\hline
V-Net \citet{b54} & 50/0 & 80.13 & 70.15 & 14.83 & 4.28 \\
\hline
DTC \citet{b1} & 50/200 & 81.17 & 70.96 & 11.51 & 2.16 \\ 
SASSnet \citet{b2} & 50/200 & 81.58 & 71.44 & 10.74 & 1.97 \\
UA-MT \citet{b3} & 50/200 & 81.21 & 71.38 & 11.96 & 3.73 \\
URPC \citet{b4} & 50/200 & 84.05 & 20.72 & 11.17 & 2.45 \\
MC-Net \citet{b7} & 50/200 & 83.16 & 73.47 & \underline{8.68} & 2.16 \\
MC-Net+ \citet{b8} & 50/200 & \underline{84.10} & \underline{74.56} & 9.99 & 2.68 \\
AC-MT \citet{b5} & 50/200 & 83.28 & 73.26 & 9.87 & \underline{1.79} \\
CAML \citet{b6} & 50/200 & 81.94 & 72.17 & 9.86 & 2.85 \\
ML-RPL \citet{b53} & 50/200 & 80.49 & 70.07 & 15.65 & 4.73 \\
Diff-CL (ours) & 50/200 & \textbf{85.45} & \textbf{75.88} & \textbf{9.05} & \textbf{2.6} \\
\hline
\hline
V-Net \citet{b54} & 100$\%$ & 86.4 & 77.43 & 6.98 & 1.79 \\
\hline
\end{tabular}}
\label{table3}
\end{table}

\begin{figure*}[!t]
\center
\includegraphics[scale=0.16]{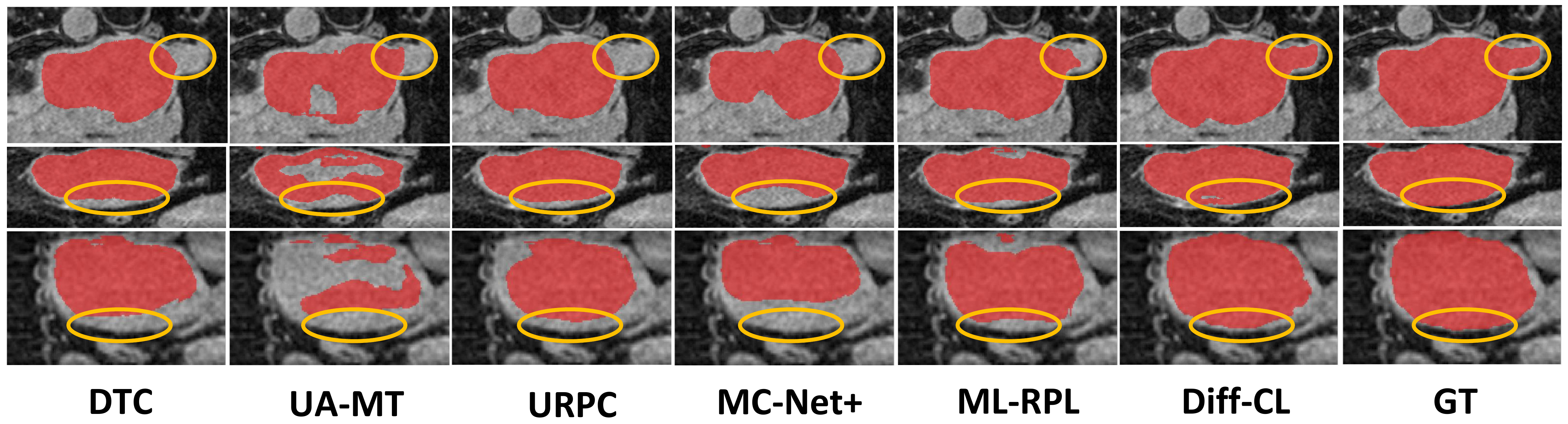} 
\caption{Results of qualitative comparison on LA dataset under 10$\%$ labeled data setting. GT represents the ground truth.}
\label{fig4}
\end{figure*}

\textbf{Implementation details.}
All the networks in our experiments are implemented on
a server with NVIDIA GeForce RTX 2080 10GB GPU, Pytorch 2.2.2+cu118, and Python 3.11. The training for DS and CS models both employ the Stochastic Gradient Descent (SGD) optimizer for 300 epochs, augmented with a weight decay of $3\times{{10}^{-5}}$ and a momentum of 0.9, while the learning rate is maintained at 0.01. Gaussian warming up function is used to control the weight $\lambda{(t)}=2.0\times{e^{-5(1-t/t_{max})^2}}$, where $t$ represents the current epoch, and $t_{max}=300$. To mitigate overfitting, we augment the training dataset with random cropping, flipping, and rotation techniques. For segmentation tasks, we utilize V-Net \citet{b54} as the backbone network of DS and CS. In LA segmentation, input patches are randomly cropped to dimensions of $112\times{112}\times{80}$ voxels, with predictions made using a sliding window strategy with strides of $18\times{18}\times{4}$ voxels. For Pancreas and BraTS segmentation, patches are randomly cropped to $96\times{96}\times{96}$ voxels, and predictions are similarly performed using a sliding window approach with strides of $16\times{16}\times{16}$ and $64\times{64}\times{64}$ voxels, respectively. 

\begin{figure*}[!t]
\centering
\includegraphics[scale=0.16]{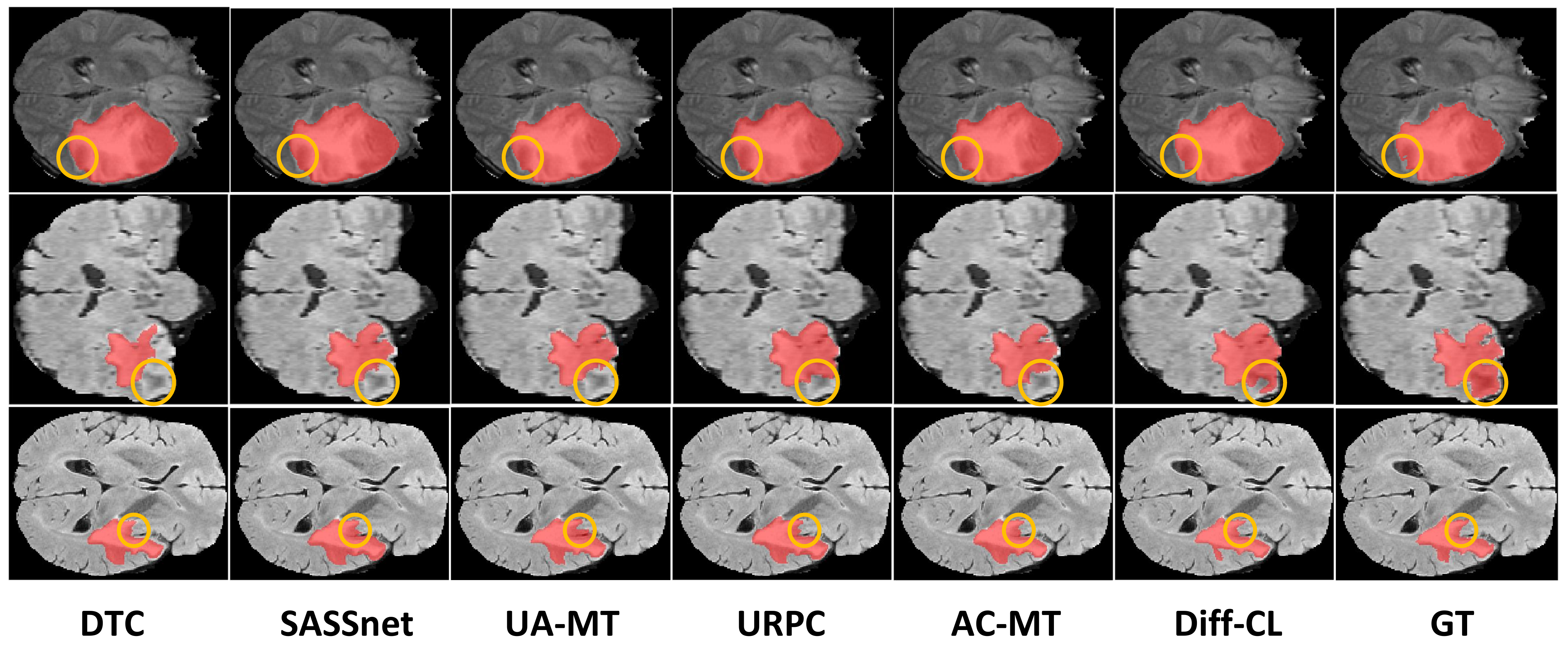}
\caption{Results of quantitative comparison on BraTS dataset. GT represents the ground truth.}
\label{fig5}
\end{figure*}

\subsection{Experimental results}
We give quantitative and qualitative analysis for segmentation results on three datasets. All experiments on LA, BraTS, and Pancreas datasets use V-Net architecture.

\subsubsection{Experiments on left atrium segmentation}
We conduct comparative experiments on LA dataset under 5$\%$ and 10$\%$ labeled data settings, respectively. 

As shown in Table \ref{table2}, Diff-CL outperforms nine other methods across three different labeled data settings. Compared nine current methods, under 5$\%$ labeled data setting, Diff-CL exceeds the highest Dice score by 1.49$\%$, Jaccard score by 2.4$\%$, and has 3.54 lower 95HD and 0.01 lower ASD than the best alternatives. Under 10$\%$, Diff-CL is 1.56$\%$ higher in Dice, 2.53$\%$ higher in Jaccard, 2.84 lower in 95HD, and 0.08 lower in ASD. These results indicate the effectiveness of Diff-CL and its near-supervised performance even with limited labeled data.

As shown in Fig. \ref{fig4}, in the highlighted areas, segmentation of Diff-CL aligns more closely with the Ground Truth, even with only 10$\%$ labeled data. Compared to methods like UA-MT and MC-Net+, Diff-CL better preserves atrial edges and details, demonstrating efficient use of limited labeled data and strong performance across various labeling ratios.

\subsubsection{BraTS segmentation dataset}
We conduct comparative experiments on BraTS dataset under 10$\%$ and 20$\%$ labeled data settings. 

As shown in Table \ref{table3}, Dice score (84.63$\%$) of Diff-CL improved by 1.02$\%$ over URPC, and its Jaccard index (75.05$\%$) increased by 1.53$\%$ under 10$\%$ labeled data setting. Dice score (85.45$\%$) of Diff-CL improved by 1.35$\%$ over URPC, and its Jaccard index (75.88$\%$) increased by 1.32$\%$ under 20$\%$ labeled data setting. However, similar to results on pancreas dataset, 95HD and ASD was at the middle level of all methods on BraTS dataset. These results indicate that Diff-CL has significant advantages in accuracy prediction, however, it needs to pay more attention to the boundary.

As shown in Fig. \ref{fig5}, the segmentation of first and third samples by Diff-CL is more accurate in detail than that by other methods in blue box. In the second sample, other methods did not segment the foreground area in blue box, and the shape of Diff-CL is more similar to ground truth.

\begin{table}[!t]
\centering
\caption{Results in quantitative comparison on Pancreas dataset.}
\renewcommand{\arraystretch}{1.2}
\setlength{\tabcolsep}{1mm}
\setlength{\tabcolsep}{5.2pt}
\begin{threeparttable}
\scalebox{0.80}{
\begin{tabular}{@{}lccccc@{}}
\hline
\multirow{2}{*}{Method} & Scans used & \multicolumn{4}{c}{Metrics}\\
\cline{2-6}
 & Labeled/Unlabeled & Dice($\%$)$\uparrow$ & Jaccard($\%$)$\uparrow$ & 95HD$\downarrow$ & ASD$\downarrow$ \\
\hline
\hline
V-Net \citet{b54} & 6/0 & 60.39 & 46.17 & 24.94 & 2.23 \\ 
\hline
DTC \citet{b1} & 6/56 & 69.01 & 54.52 & 20.99 & 2.33 \\ 
SASSnet \citet{b2} & 6/56 & 72.28 & 58.06 & 14.30 & 2.45 \\
UA-MT \citet{b3} & 6/56 & 71.26 & 56.15 & 22.01 & 7.36 \\
URPC \citet{b4} & 6/56 & 72.66 & 18.99 & 22.63 & 6.36 \\
MC-Net \citet{b7} & 6/56 & 69.96 & 55.57 & 23.90 & 1.56 \\
MC-Net+ \citet{b8} & 6/56 & 63.31 & 49.15 & 31.99 & 2.09 \\
AC-MT \citet{b5} & 6/56 & 73.00 & 58.93 & 18.36& 1.91 \\
CAML \citet{b6} & 6/56 & 70.33 & 55.85 & 20.60 & \underline{1.71} \\
ML-RPL \citet{b53} & 6/56 & \underline{77.95} & \underline{64.53} & \underline{8.77} & 2.29 \\
Diff-CL (ours) & 6/56 & \textbf{78.21} & \textbf{64.8} & \textbf{14.11} & \textbf{3.26} \\
\hline
V-Net \citet{b54} & 12/0 & 71.52 & 57.68 & 18.12 & 5.41 \\ 
\hline
DTC \citet{b1} & 12/50 & 73.55 & 59.90 & 13.55 & 1.59 \\ 
SASSnet \citet{b2} & 12/50 & 77.43 & 64.18 & 11.78 & 1.53 \\
UA-MT \citet{b3} & 12/50 & 76.66 & 63.09 & 11.85 & 3.53 \\
URPC \citet{b4} & 12/50 & 75.22 & 22.75 & 13.86 & 4.06 \\
MC-Net \citet{b7} & 12/50 & 74.80 & 60.57 & 19.18 & 4.64 \\
MC-Net+ \citet{b8} & 12/50 & 73.71 & 60.34 & 13.93 & 4.00 \\
AC-MT \citet{b5} & 12/50 & 78.86 & 66.02 & \underline{7.98} & 1.47 \\
CAML \citet{b6} & 12/50 & 74.96 & 61.81 & 14.60 & \underline{1.29} \\
ML-RPL \citet{b53} & 12/50 & \underline{80.29} & \underline{67.53} & 9.50 & 2.21 \\
Diff-CL (ours) & 12/50 & \textbf{81.27} & \textbf{68.87} & \textbf{9.32} & \textbf{2.16} \\
\hline
\hline
V-Net \citet{b54} & 100$\%$ & 82.60 & 70.81 & 5.61 & 1.33  \\
\hline
\end{tabular}}
\label{table4}
\end{threeparttable} 
\end{table}

\begin{figure*}[!t]
\centering
\includegraphics[scale=0.12]{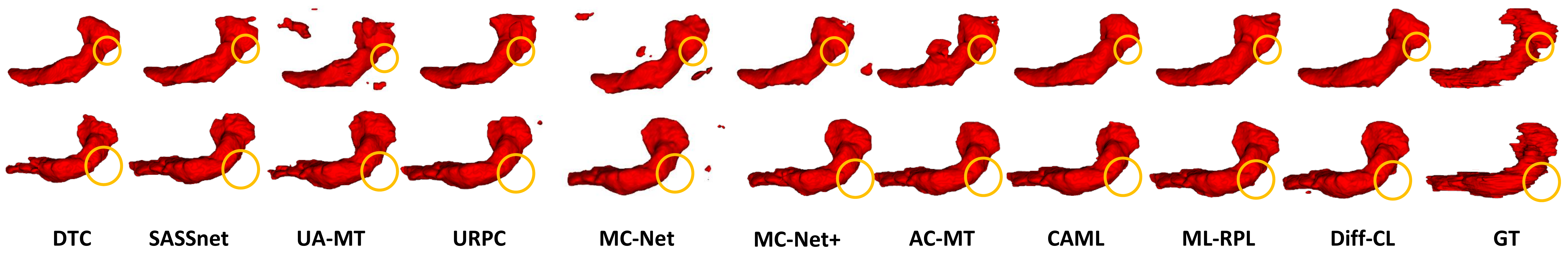}
\caption{Results of qualitative comparison on Pancreas dataset under 10$\%$ labeled data setting. GT represents the ground truth.}
\label{fig6}
\end{figure*}

\subsubsection{Experiments on NIH pancreas segmentation}
We conduct comparative experiments on the pancreas dataset under 10$\%$ and 20$\%$ labeled data settings. 

In Table \ref{table4}, under 10$\%$ labeled data setting, Diff-CL achieves a Dice score of 78.21$\%$ and a Jaccard index to 64.80$\%$, which is 0.26$\%$ and  0.27$\%$ higher than the best of 77.95$\%$ and 64.53$\%$ by ML-RPL, respectively. Under 20$\%$ labeled data setting, Diff-CL achieves a Dice score of 81.27$\%$ and a Jaccard index to 68.87$\%$, which is 0.98$\%$ and  1.34$\%$ significantly higher than the best of 80.29$\%$ and 67.53$\%$ by ML-RPL, respectively, which demonstrate superior segmentation accuracy. However, 95HD and ASD are not the best of all these methods, so Diff-CL must further improve the boundary precision on pancreas dataset.

As shown in Fig. \ref{fig6}, in the first row, within the blue circle, Diff-CL identifies the furrow, while other methods fail to detect it. In the second row, within the blue circle, Diff-CL successfully segments a synapse that closely resembles the ground truth (GT), whereas the other methods miss this segment altogether. This highlights the superior capability of Diff-CL in accurately capturing fine structures that other methods overlook.

\subsection{Ablation study}

We have done ablation studies to verify the effectiveness of Diff-CL. Table \ref{table6} presents results of ablation studies on three datasets: Pancreas and LA. We compare different combinations of loss functions and evaluate their impact on various metrics: Dice, Jaccard, 95HD, and ASD. Using only supervised loss including $L^c_{s}$ and $L^d_{s}$ provides a solid foundation for segmentation accuracy but leaves room for improvement in boundary approximation and segmentation smoothness. Adding cross-pseudo-supervision losses $L^c_{p}$ and $L^d_p$ and the high-frequency mamba (HFM) module significantly improves segmentation accuracy and overlap (Dice and Jaccard). It also enhances boundary approximation and smoothness, as evidenced by lower 95HD and ASD values. Including inter-sample contrastive loss yields the highest accuracy and best overlap (Dice and Jaccard) among all setups. It also substantially reduces the worst-case boundary error (95HD) and improves contour smoothness (ASD). This research shows the effectiveness of Diff-CL.

\begin{table}[!t]
\centering
\caption{Ablation studies on Pancreas and LA datasets under 5$\%$, 10$\%$ and 20$\%$ labeled data setting respectively.}
\renewcommand{\arraystretch}{1.0}
\setlength{\tabcolsep}{8.0pt}
\begin{tabular}{@{}cccccccc@{}}
\hline
\multicolumn{4}{c}{Methods} & \multicolumn{4}{c}{Metrics}\\
\hline
 $L^c_s + L^d_{s}$ & $L^c_p+L^d_p$ & HFM & $L_{cl}$ & Dice($\%$) & Jaccard($\%$) & 95HD & ASD \\
\hline
\multicolumn{8}{c}{Pancreas Dataset} \\
\hline
$\checkmark$ & & &  & 71.52 & 57.68 & 18.12 & 5.41 \\ 
$\checkmark$ & $\checkmark$ & & & 80.36 & 67.81 & 8.85 & 2.00 \\
$\checkmark$ & $\checkmark$ & $\checkmark$ & & 80.85 & 68.32 & 9.75 & 2.62 \\
$\checkmark$ & $\checkmark$ & $\checkmark$ & $\checkmark$ & 81.27 & 68.87 & 9.32 & 2.16 \\
\hline
\multicolumn{8}{c}{LA Dataset} \\
\hline
$\checkmark$ & & &  & 84.81 & 76.48 & 8.84 & 2.14 \\ 
$\checkmark$ & $\checkmark$ & & & 86.54 & 77.70 & 6.02 & 2.26 \\
$\checkmark$ & $\checkmark$ & $\checkmark$ & & 88.17 & 79.23 & 8.23 & 2.8 \\
$\checkmark$ & $\checkmark$ & $\checkmark$ & $\checkmark$ & 89.02 & 80.34 & 6.32 & 2.16\\
\hline
\end{tabular}
\label{table6}
\end{table}

\section{Conclusion}
\label{section 6}
This paper proposes a new semi-supervised medical image segmentation framework from the distribution perspective. On the one hand, we propose a crosse-pseudo-supervision learning mechanism of diffusion and convolution segmentation networks. Combining diffusion models with convolutional neural networks enables the former to learn data distribution while the latter corrects fine details. Consider capturing complete high-frequency details by CNNs requires substantial computational resources and is susceptible to local noise, we design a 3D high-frequency Mamba module to learn high-frequency details in medical images. On the other hand, given that both labeled and unlabeled data come from the same distribution, we utilize contrastive learning to verify our idea that regions in unlabeled data similar to overall class semantics to labeled data are likely to belong to the same class, while regions with minimal similarity are less likely to. The superior segmentation performance on the NIH pancreas, left atrium, and brain tumor datasets demonstrate the effectiveness of Diff-CL. 

\section*{Declaration of interests}
The authors declare that they have no known competing financial interests or personal relationships that could have appeared to influence the work reported in this paper.
\section*{Data availability}
Data will be available upon request.

\section*{Acknowledgments}
This work was supported by National Natural Science Foundation of China Grants (No.12101426 and No.12426308), Beijing Outstanding Young Scientist Program (No. JWZQ20240101027), and Beijing Natural Science Foundation Grants (No.Z210003 and No.4254093).




\end{document}